\documentclass[runningheads]{llncs}

\usepackage{eccv}

\usepackage{eccvabbrv}

\usepackage{graphicx}
\usepackage{booktabs}
\usepackage[para]{footmisc}

\usepackage[accsupp]{axessibility}  

\usepackage[pagebackref,breaklinks,colorlinks,citecolor=eccvblue]{hyperref}

\usepackage{orcidlink}

\usepackage{colortbl}
\usepackage{amssymb}
\usepackage{pifont}
\newcommand{\cmark}{\ding{51}}
\newcommand{\xmark}{\ding{55}}
\newcommand{\methodname}{Uni3DR$^2$\xspace}
\newcommand{\snowflake}{\includegraphics[width=10px]
{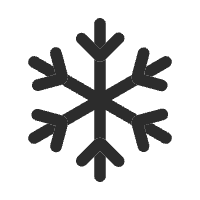}}

\begin{document}

\title{Unified Scene Representation and Reconstruction \\ for 3D Large Language Models} 

\titlerunning{\methodname}

\author{Tao Chu\inst{1,2,*} \and
Pan Zhang\inst{2} \and
Xiaoyi Dong\inst{2} \and
Yuhang Zang\inst{2} \and
Qiong Liu\inst{1, \dag} \and \\
Jiaqi Wang\inst{2}}

\authorrunning{T. Chu et al.}

\institute{South China University of Technology \\
\email{liuqiong@scut.edu.cn} \and 
Shanghai AI Laboratory  \\
\email{\{chutao,zhangpan,dongxiaoyi,wangjiaqi\}@pjlab.org.cn}}

\maketitle
\renewcommand{\thefootnote}{\fnsymbol{footnote}}
\footnotetext[1]{\hspace{-0.5em}Intern at Shanghai AI Laboratory.}
\footnotetext[2]{\hspace{-0.5em}Corresponding author.}

\begin{abstract}

Enabling Large Language Models (LLMs) to interact with 3D environments is challenging. Existing approaches extract point clouds either from ground truth (GT) geometry or 3D scenes reconstructed by auxiliary models. Text-image aligned 2D features from CLIP are then lifted to point clouds, which serve as inputs for LLMs. However, this solution lacks the establishment of 3D point-to-point connections, leading to a deficiency of spatial structure information. 
Concurrently, the absence of integration and unification between the geometric and semantic representations of the scene culminates in a diminished level of 3D scene understanding.
In this paper, we demonstrate the importance of having a unified scene representation and reconstruction framework, which is essential for LLMs in 3D scenes.
Specifically, we introduce \methodname
extracts 3D geometric and semantic aware representation features via the frozen pre-trained 2D foundation models (e.g., CLIP and SAM) and a multi-scale aggregate 3D decoder.
Our learned 3D representations not only contribute to the reconstruction process but also provide valuable knowledge for LLMs.
Experimental results validate that our \methodname yields convincing gains over the baseline on the 3D reconstruction dataset ScanNet (increasing F-Score by +1.8\%). When applied to LLMs, our \methodname-LLM exhibits superior performance over the baseline on the 3D vision-language understanding dataset ScanQA (increasing BLEU-1 by +4.0\% and +4.2\% on the val set and test set, respectively). Furthermore, it outperforms the state-of-the-art method that uses additional GT point clouds on both ScanQA and 3DMV-VQA.
\keywords{3D Reconstruction \and 3D Representation \and Large Language Models}

\end{abstract}

\section{Introduction}
\label{sec:intro}

\begin{figure}
    \centering
    \includegraphics[width=.7\linewidth]{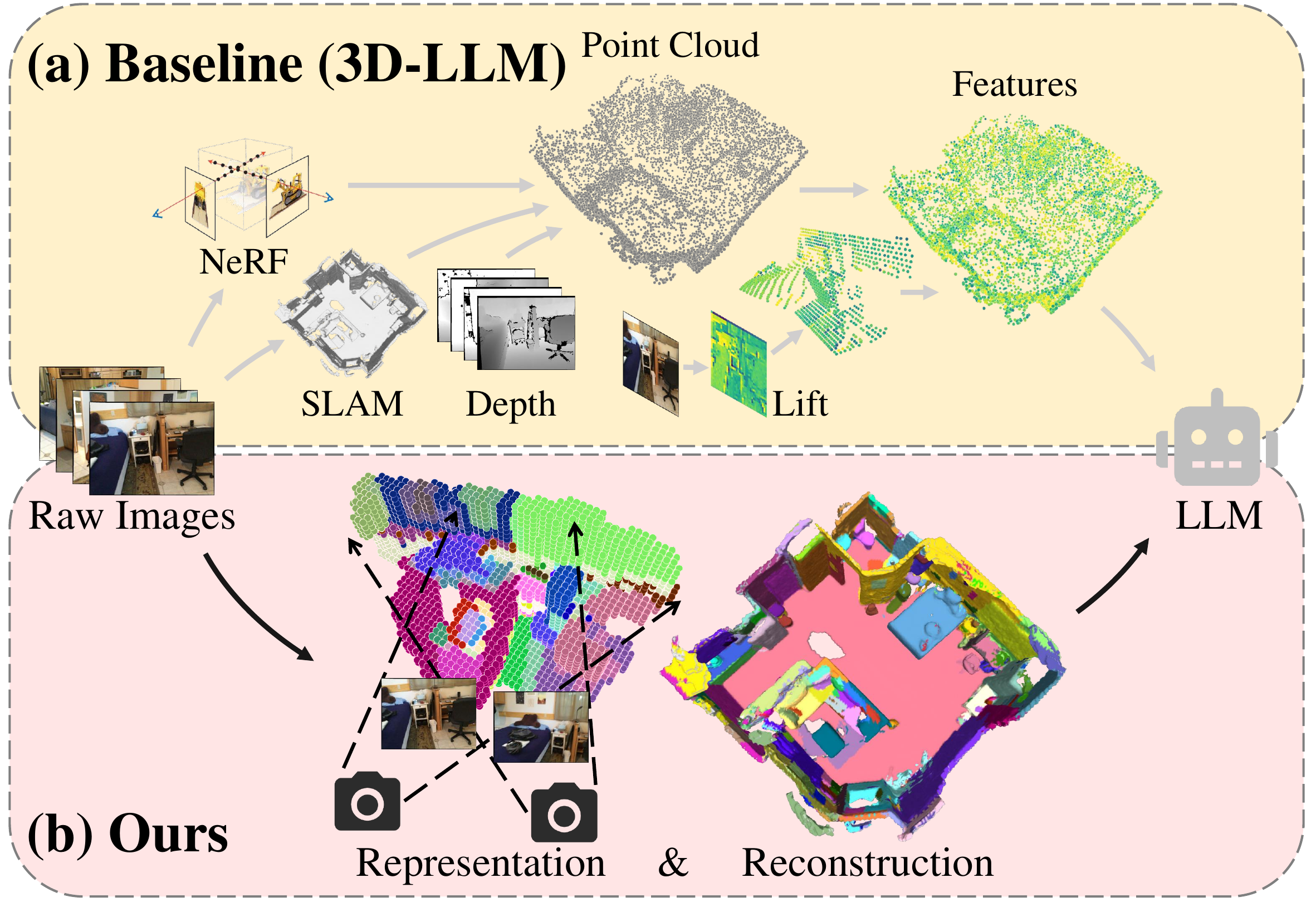}
    \caption{
    \textbf{Comparison between the representation for 3D LLMs.} 
    \textbf{(a)} The previous baseline~\cite{hong20233d} is isolated and complex, which requires extra NeRF, SLAM models, or depth to extract the point cloud and then lifts features to the 3D representations.
    \textbf{(b)} By contrast, our \methodname is unified and neat. We unified learn geometric and semantically rich volumetric representation with high-quality reconstruction as LLM inputs. Our learned representation and reconstruction significantly enhance the LLM's performance in 3D environments.
    }
    \vspace{-15pt}
    \label{fig:teaser}
\end{figure}

Recent Large Language Models (LLMs)~\cite{chatgpt2022,touvron2023llama,chowdhery2022palm,openai2023gpt4} exhibit remarkable proficiency in processing both 1D text and 2D images.
A key factor contributing to their notable success is the LLMs' capability to learn comprehensive representations from extensive sets of language and visual data. However, LLMs face challenges with representation learning in 3D environments, which are unstructured and inherently complex. Directly using a sequence of 2D image inputs for 2D LLMs fails to yield satisfactory results in the 3D world~\cite{hong20233d}. Therefore, learning high fidelity and accuracy of 3D representations could enable more intuitive and versatile AI systems that understand and interact with the physical 3D world in a more human-like manner.

A recent advancement 3D-LLM~\cite{hong20233d} marks the inaugural attempt to utilize point clouds as representations for 3D LLMs, as illustrated in \cref{fig:teaser} (\textbf{a}). However, 3D-LLM generates representations offline through an isolated and complex process, lifting 2D features to point clouds extracted from input depth, ground truth (GT) geometry, or 3D scenes reconstructed by auxiliary models.
While using GT point clouds as input can yield commendable performance for 3D-LLM, the acquisition of such GT data is both expensive and challenging. Conversely, depending on the geometry predicted by the reconstruction models as input to 3D-LLM has demonstrated subpar performance, as discussed in \cref{sec:ablation}.
This performance discrepancy arises because 3D-LLM relies solely on point clouds to provide spatial positions without establishing 3D point-to-point connections or fusing 3D geometric and semantic information.
Consequently, the 3D representations exhibit sparse scene connectivity and a lack of spatial structural information, causing 3D-LLM's performance to be notably affected by the quality of the scene geometry. 
These limitations highlight the necessity for further advancements in 3D representation design, which can effectively bridge the gap between LLMs and complex 3D environments.

To this end, we present \methodname (\cref{fig:teaser} (\textbf{b})), a neat and unified scene representation and reconstruction module for LLM.
And \methodname-LLM addresses the substantial impact of predicted geometry on 3D-LLM, achieving results that surpass the performance of 3D-LLM which uses GT point clouds as inputs.
Specifically, \methodname comprises a 2D encoder and a 3D decoder for 3D representations, and a reconstruction module.
Unlike the previous reconstruction methods that utilize trainable encoders solely for geometric features, our 2D encoder is combined with two frozen vision foundation backbones. Notably, one backbone is derived from SAM~\cite{kirillov2023segment}, pretrained on extensive sets of human-verified object masks, for extracting object-level information. The other backbone is sourced from CLIP~\cite{radford2021learning}, pretrained on massive image-text data, to capture semantically rich features.
Then a 3D decoder equipped with multi-scale Gated Recurrent Unit (GRU) fusion is introduced for the generation of 3D geometric and semantically rich representations from 2D features.
Subsequently, a lightweight reconstruction module leverages these 3D representations to predict precise geometry results. Finally, the combined representation and reconstruction outcomes serve as inputs for LLMs to address 3D vision-language tasks. In comparison to previous methods, our unified scene representation and reconstruction module excels in capturing rich semantics and object details within 3D spaces, steering clear of the use of expensive GT point clouds or the performance limitations associated with relying solely on predicted point clouds.

Extensive experiments show that our proposed method outperforms prior competitive approaches on available datasets that include video inputs. In terms of reconstruction, our \methodname attains a remarkable 1.8\% improvement in F-Score compared to the baseline on ScanNet. Regarding 3D vision-language tasks, in comparison to the baseline 3D-LLM~\cite{hong20233d} with predicted geometry,
our \methodname-LLM also exhibits significant 4.0\% and 4.2\% increases in BLEU-1 on the \textit{val} set and \textit{test} set of ScanQA, respectively. Moreover, it outperforms the state-of-the-art which uses additional GT point clouds. Specifically, our \methodname-LLM achieves a notable gain of $3.4\%$ in overall accuracy on 3DMV-VQA and a commendable improvement of $1.4\%$ in BLEU-1 on the \textit{test} set of ScanQA, respectively.

\section{Related Work}
\noindent \textbf{3D Vision and Language} plays a pivotal role in enabling machines to comprehend 3D environments in applications like robotics~\cite{ha2022semantic,thomason2022language} and embodied AI~\cite{szot2021habitat,khandelwal2022simple}. Several recent 3D tasks~\cite{azuma2022scanqa,hong20233d2,chen2021scan2cap,chen2020scanrefer,achlioptas2020referit3d} evaluate the visual and language understanding abilities of 3D models, such as question answering~\cite{azuma2022scanqa,ye20223d,hong20233d2}, dense captioning~\cite{chen2021scan2cap}, and grounding~\cite{chen2020scanrefer,achlioptas2020referit3d}.
State-of-the-art algorithms commonly use graph representation~\cite{huang2021text,feng2021free} or multimodal Transformer~\cite{cai20223djcg,chen2023unit3d,wu2023eda,yang2021sat,zhu20233d}.
Recent work 3D-LLM~\cite{hong20233d} combines 3D representations and large language models (LLMs)~\cite{li2023blip2} to solve 3D vision-language tasks.
Specifically, 3D-LLM extracts 3D representations from 2D images via extra depth map, SLAM~\cite{jatavallabhula2019gradslam}, or NeRF model~\cite{sun2022direct}.
Different from 3D-LLM, our \methodname uses the volumetric representation and reconstruction from a sequence of images, which is more accurate and general in handling different scenes for LLMs.
Our \methodname primarily focuses on feature representation and reconstruction, which is orthogonal to previous methods~\cite{huang2023embodied} and can be integrated together.

\noindent \textbf{3D Reconstruction} aims to reconstruct 3D scenes from input images, which main contains depth-based methods~\cite{sayed2022simplerecon, choe2021volumefusion, rich20213dvnet} and volumetric methods~\cite{saito2019pifu, sun2021neuralrecon}. 
SimpleRecon~\cite{sayed2022simplerecon} is proposed to input metadata into the feature volume and then follow a 2D encoder-decoder to predict depth.
VolumeFusion~\cite{choe2021volumefusion} utilizes differential depth fusion with proposed PosedConv to optimize multi-view depth maps.
3DVNet~\cite{rich20213dvnet} constructs a multi-scale volumetric scene encoding from a set of estimated depth map inputs and then predicts a residual for these depth maps.
PIFu~\cite{saito2019pifu} predicts the continuous inside/outside probability field to obtain the 3D occupancy.
NeuralRecon~\cite{sun2021neuralrecon} is proposed to predict a discrete TSDF volume using GRU fusion.
Depth-based reconstruction methods are prone to disturbances in depth values, making it challenging to obtain stable reconstruction results. Volumetric methods demonstrate stronger stability and can accommodate more feature information in 3D space. However, previous volumetric reconstruction approaches~\cite{saito2019pifu, sun2021neuralrecon} usually rely on small, learnable image encoders, which limited their ability to provide semantically rich 3D representations.
Different from previous methods, we use the frozen encoders from segmentation-anything (SAM)~\cite{kirillov2023segment} and CLIP~\cite{radford2021learning} to extract semantically rich 3D representations.

\noindent \textbf{2D Vision Foundation Models} are built upon the pre-training technique that aims to train a general model using massive data and can be fine-tuned easily in different downstream tasks. For example, CLIP model~\cite{radford2021learning} is pre-trained on large amounts of image-text pairs with contrastive loss. SAM~\cite{kirillov2023segment} model is another example that is pre-trained on billion-level images with perception prompts such as points, boxes, and segmentation masks. These vision foundation models contain rich semantic information for downstream applications.
Our \methodname uses these foundation models as image encoders, extracting both rich geometric and semantic features.


\noindent \textbf{Large Language Models} represent a groundbreaking advancement in Natural Language Processing, exhibiting exceptional performance across diverse tasks. At the forefront of this technology is OpenAI's GPT series \cite{radford2018improving,radford2019language,openai2023gpt4}. These models have set new standards in language understanding and generation, with a profound impact on various applications. In addition
, other notable LLMs have emerged, such as PaLm \cite{chowdhery2022palm}, OPT \cite{zhang2022opt} and LLaMa \cite{touvron2023llama}. Notably, the field has seen recent strides in the development of Multimodal LLMs, such as GPT-4~\cite{openai2023gpt4}, BLIP2~\cite{li2023blip2}, PaLmE~\cite{driess2023palme} and 3D-capable models 3D-LLM~\cite{hong20233d}. These advancements mark a pivotal moment, extending the capabilities of LLMs beyond language processing to seamlessly integrate and interpret both textual and visual information.
Our \methodname-LLM is built upon BLIP2~\cite{li2023blip2} with a unified representation and reconstruction framework for 3D environments.

\section{Methodology}

\begin{figure*}[tb]
    \centering
    \includegraphics[width=\linewidth]{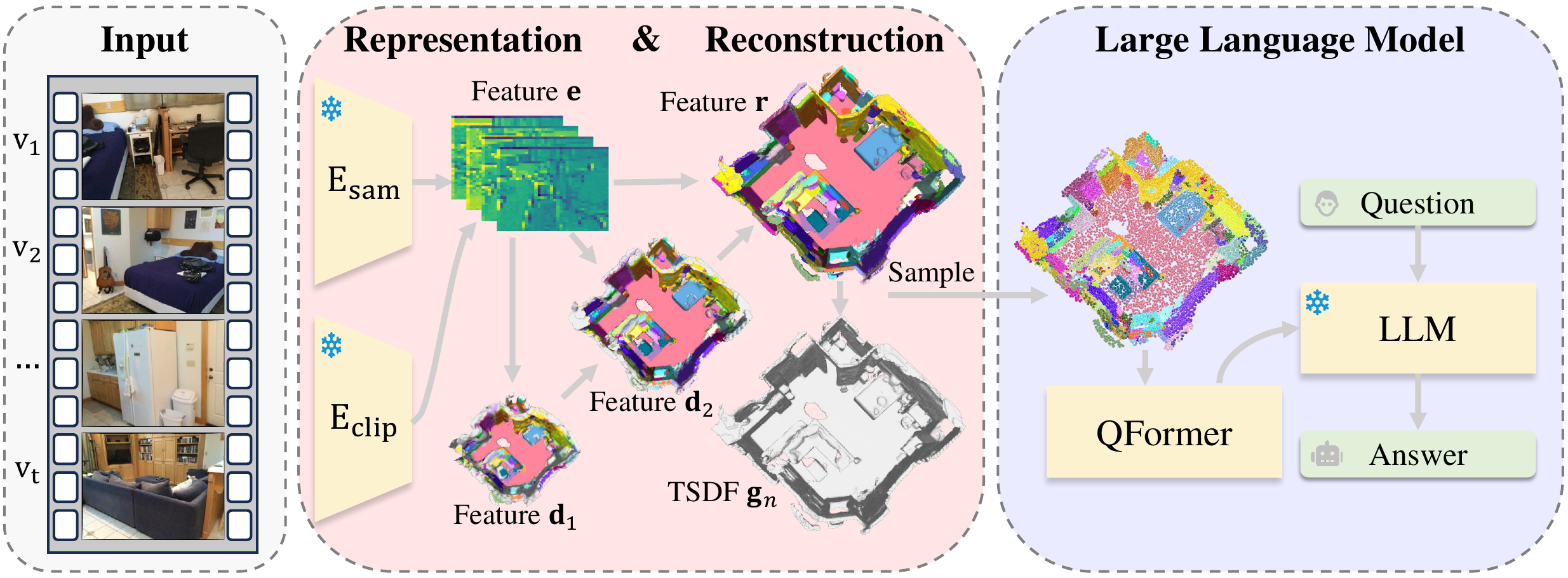}
    \caption{
    \small
    \textbf{Overview of our \methodname-LLM framework.} Given video inputs, \methodname-LLM employs \textbf{(1)} a frozen encoder integrating SAM~\cite{kirillov2023segment} and CLIP~\cite{radford2021learning} image encoders, followed by a decoder for \textbf{3D representations}, \textbf{(2)} a light-weight module focus on \textbf{3D reconstruction}, and \textbf{(3)} an LLM integrated with QFormer~\cite{li2023blip2} for \textbf{3D vision-language understanding}.
    (\snowflake: frozen.).
    }
    \label{fig:framework}
    \vspace{-10pt}
\end{figure*}

This section describes the framework of our \methodname-LLM --- a unified model understanding 3D scenes from 2D video inputs.
\cref{fig:framework} illustrates the framework of \methodname-LLM, which comprises a unified module for predicting 3D representations $\mathbf{r}$ and reconstruction results $\mathbf{g}_n$, and a Large Language Model for 3D vision-language tasks.
Unlike the previous method~\cite{hong20233d}, which lifts extracted 2D features into expensive GT point clouds or predicted geometry for LLMs, resulting in generated representation with sparse scene connectivity and a lack of spatial structural information, our proposed \methodname aims to establish 3D geometric and semantically rich representations for both reconstruction and LLMs.

\vspace{-5pt}
\subsection{Unified Representation and Reconstruction (\methodname)}
\vspace{-5pt}
Given a video $\mathbf{V} = \{ \mathbf{v}_{1}, \mathbf{v}_{2}, \ldots, \mathbf{v}_{t} \}$, our \methodname aim to map it to 3D representation features $\mathbf{r}$ and 3D reconstruction results $\mathbf{g}_n$. Here, $t$ serves as a temporal sequence identifier, and $\mathbf{v} \in \mathbb{R}^{H\times W \times 3} $ represents an image frame, with $H$ and $W$ denoting the image height and width, respectively.
Specifically, the video frames $\mathbf{V}$ are input to the 2D strong encoder $\mathbf{E}$ for object-level and semantically rich features $\mathbf{e}$, and $\mathbf{e}$ are then fed into the 3D decoder $\mathbf{D}$ for 3D point-to-point connection and feature fusion. The decoder $\mathbf{D}$ generates 3D geometric and semantically rich representations $\mathbf{r}$. Finally, $\mathbf{r}$ is used to predict 3D geometry $\mathbf{g}_n$ with a light-weight reconstruction module.

\noindent \textbf{Encoder.}
Our encoder $\mathbf{E}$ is combined of two foundation pre-trained visual encoders: $\mathbf{E}_{\text{sam}}$ from SAM~\cite{kirillov2023segment} and $\mathbf{E}_{\text{clip}}$ from CLIP~\cite{radford2021learning}. Since $\mathbf{E}_{\text{sam}}$ and $\mathbf{E}_{\text{clip}}$ are pre-trained on the massive object-level masks and image-text corpus, respectively, we first use them to extract object-level and semantically rich features:
\vspace{-10px}
\begin{equation}
\vspace{-5px}
\begin{aligned}
\mathbf{e}_{\text{sam}} = \mathbf{E}_{\text{sam}}(\mathbf{V}) \in \mathbb{R}^{B\times C_{0}\times H_{0} \times W_{0}}, \\
\mathbf{e}_{\text{clip}} = \mathbf{E}_{\text{clip}}(\mathbf{V}) \in \mathbb{R}^{B \times C_{1} \times H_{1} \times W_{1}},\\
\end{aligned}
\end{equation}
where symbols $B$, $C$, $H$, and $W$ refer to batch size, output channels, height, and width, respectively.
And then, concatenation ($\texttt{concat}$) and reshape ($\texttt{reshape}$) operations are used to get final 2D encoder features $\mathbf{e} = \texttt{concat}(\texttt{reshape}\left(\mathbf{e}_{\text{sam}}\right), $\texttt{\\reshape}$\left({\mathbf{e}_{\text{clip}}}\right)) \in \mathbb{R}^{B \times C_{2} \times H_{2} \times W_{2}}.$ Here the output channel $C_{2} = C_{0} + C_{1}$.

Unlike previous 3D reconstruction methods~\cite{im2019dpsnet, sun2021neuralrecon} where the encoders are trained for reconstruction, our method introduces the frozen foundation encoders in reconstruction for the first time. These encoders can naturally extract 2D features with rich information, enabling them to effectively capture both geometric and semantic details essential for 3D representations.




\noindent \textbf{Decoder.} We use a 3D decoder $\mathbf{D}$ to fuse the 2D features $\mathbf{e}$ into 3D representations $\mathbf{r}$. The decoder $\mathbf{D}$ builds the relationship of voxels in 3D space with multi-scale and coarse-to-fine strategies. Specifically, the encoder feature $\mathbf{e}$ is input into $n$ levels of convolution operators. This process yields the multi-scale features, denoted as $\mathbf{e}^{in} = \{\mathbf{e}^{in}_{1}, \mathbf{e}^{in}_{2}, \ldots, \mathbf{e}^{in}_{n} \}$. For the $i$-th level, 
\vspace{-5px}
\begin{equation}
\vspace{-5px}
  \mathbf{e}_i^{in} = \texttt{conv}_i(\mathbf{e}),
\end{equation}
where convolution $\texttt{conv}_i$ is used to adjust the inputs of the $i$-th level in decoder, whose stride is $2^{(n-i)}$. And 3D decoder feature $\mathbf{d}_{i}$ of level $i$ is derived as follows:
\vspace{-5px}
\begin{equation}\label{eq:decoder}
\vspace{-3px}
\mathbf{d}_i =
\begin{cases} 
\mathbf{m}_{i}(\texttt{backproj}(\mathbf{e}_1^{in})), & \text{if }\ i = 1, \\
\mathbf{m}_{i}(\mathbf{d}_{i-1}^\texttt{filter}, \texttt{backproj}(\mathbf{e}_i^{in})), & \text{otherwise},
\end{cases}
\end{equation}
where $\texttt{backproj}$ is a backprojection function which lifts 2D features to the corresponding camera ray within 3D space with the help of camera pose. We adopt GRU Fusion~\cite{sun2021neuralrecon} following SPVConv~\cite{tang2020searching} as our 3D module $\mathbf{m}_{i}$, which is used to fuse features of each voxel in 3D space at level $i$. The symbol $\mathbf{d}_{i}^\texttt{filter} = \mathbf{o}_i \times \mathbf{d}_{i}$ represents the process of filtering  $\mathbf{d}_{i}$ using the 3D occupancy $\mathbf{o}_i$, as further detailed in Equation~\eqref{eq:reconstruction}.
Finally, the 3D representations $\mathbf{r}=\mathbf{d}_n$ are obtained from the last level of the 3D decoder $\mathbf{D}$.

The previous 3D LLM method~\cite{hong20233d} directly lifts 2D features to GT/predicted point clouds as 3D representations, resulting in sparse scene connectivity and a lack of spatial structural information. In contrast, our 3D decoder tackles these issues by establishing 3D point-to-point connections, fusing both geometric and semantic information into 3D representations. Consequently, it enhances both reconstruction and LLMs.



\noindent \textbf{Reconstruction.} After obtaining the 3D representations $\mathbf{r}$, the subsequent steps involve acquiring 3D reconstruction results $\mathbf{g}_{n}$.
The symbol $\mathbf{g}_{n}$ denotes to the Truncated Signed Distance Function (TSDF).
For reconstruction, we predict geometry in each level to filter the free regions in 3D space due to limited GPU memory. The reconstruction results at level $i$ are denoted as $\mathbf{g}_i$ via Equation~\eqref{eq:reconstruction}:
\vspace{-8px}
\begin{equation}\label{eq:reconstruction}
\vspace{-8px}
\begin{aligned}
\mathbf{g}_i &= \mathbf{o}_i \times \mathbf{h}_i^{tsdf}(\mathbf{d}_{i}), \\
\mathbf{o}_i &= \mathbf{h}_i^{occ}(\mathbf{d}_{i}),
\end{aligned}
\end{equation}
where $\mathbf{h}_i^{tsdf}$ and $\mathbf{h}_i^{occ}$ represent the TSDF head and 3D occupancy head, respectively. Both $\mathbf{h}_i^{tsdf}$ and $\mathbf{h}_i^{occ}$ are composed of linear layers. The predicted geometry at level $n$ is denoted as $\mathbf{g}_n$, which is the reconstruction result of our model.

Our \methodname first highlights the significance of unified representation and reconstruction for LLMs within 3D scenes. Unlike the previous method~\cite{hong20233d} which relies on expensive GT point clouds or faces performance limitations when lifting features to predicted point clouds, our method excels by simultaneously predicting geometry and delivering 3D geometric and semantically rich representations.


\subsection{\methodname-LLM}
After getting representation features $\mathbf{r}$ and reconstruction results $\mathbf{g}_n$, we execute a sampling operation guided by $\mathbf{g}_n$ to prepare 3D representations for LLM. The sampled 3D representation features $\mathbf{r}$, denoted as $\mathbf{r_{s}}$, are obtained using the random sampling function $\texttt{randsamp}$:
\vspace{-8px}
\begin{equation}
\vspace{-8px}
    \mathbf{r_{s}} = \texttt{randsamp}(\mathbf{r} \times (|\mathbf{g}_n| < \delta) , N).
\end{equation}

Here, $\mathbf{r_{s}} \in \mathbb{R}^{B\times N \times C_{3}}$, where $N$ represents the fixed point number for the random sampling function, and $C_{3}$ is the channel number. The hyper-parameter $\delta$ denotes the threshold distance from the TSDF surface, guiding the selection of the sampled 3D representation features $\mathbf{r_{s}}$.

Subsequently, we combine the sampled 3D representations $\mathbf{r_{s}}$ with the 3D point coordinates $\mathbf{c} \in \mathbb{R}^{B\times N \times 3}$ to serve as the inputs for the LLM. The coordinates $\mathbf{c}$ refer to the points in $\mathbf{r_{s}}$. To enhance spatial understanding and capture location information, our position embeddings $\mathbf{p}$ are derived by concatenating the position embeddings along each axis dimension:
\vspace{-8px}
\begin{equation}
\vspace{-8px}
\mathbf{p} = \texttt{concat}\{
\texttt{L}_x(\mathbf{c}_{x}),
\texttt{L}_y(\mathbf{c}_{y}), \texttt{L}_z(\mathbf{c}_{z})
\}.
\end{equation}

Give the axis dimension $j \in \{x, y, z\}$, the symbol $\mathbf{c}_j$ denotes the values along axis $j$, and $\texttt{L}_j$ refers to the embedding layer of axis $j$. The concatenation operation $\texttt{concat}$ is performed along the channel dimension.

After that, the 3D representation features and position embeddings are fused through the expression:
\vspace{-12px}
\begin{equation}
\vspace{-8px}
    \mathbf{r_{llm}} = \texttt{proj}_{r}(\mathbf{r_{s}}) + \texttt{proj}_{p}(\mathbf{p}),
\end{equation}
where $\texttt{proj}$ refers to linear layers that adjust channel numbers. The results $\mathbf{r_{llm}}$ constitutes the final feature inputs for the LLM.

Drawing inspiration from BLIP2~\cite{li2023blip2}, we first feed our input features $\mathbf{r_{llm}}$ into the QFormer model. Subsequently, the obtained features undergo processing through LLM for dialogue generation. The 3D visual features for LLM are computed as follows:
\vspace{-6px}
\begin{equation}
\vspace{-6px}
    \mathbf{l_{llm}} = \texttt{QFormer}(\mathbf{r_{llm}}).
\end{equation}

Following this, we keep the 3D visual features $\mathbf{l_{llm}}$ fixed and introduce questions $\mathbf{q}$ to LLM, generating corresponding answers $\mathbf{a}$ through the expression:
\vspace{-8px}
\begin{equation}
\vspace{-6px}
    \mathbf{a} = \texttt{LLM}(\mathbf{q}, \mathbf{l_{llm}}).
\end{equation}

In summary, our \methodname-LLM uniformly predicts both scene representations and reconstructions, feeding these 3D geometric and semantically rich representations into LLM for engaging dialogue with 3D scenes.

\subsection{Loss}
The loss function for \methodname-LLM is a combination of two crucial components: the 3D reconstruction loss $\mathcal{L}_{rec}$ and the LLM loss $\mathcal{L}_{llm}$, represented as:
\vspace{-8px}
\begin{equation}
\vspace{-8px}
    \mathcal{L} = \mathcal{L}_{rec} + \mathcal{L}_{llm}.
\end{equation}

The reconstruction loss $\mathcal{L}_{rec}$ is designed to ensure accurate 3D reconstruction across multiple levels and is computed as follows:
\vspace{-8px}
\begin{equation}
\vspace{-8px}
    \mathcal{L}_{rec} = \sum_{i=1}^n \texttt{BCE}(\mathbf{o}_i, \mathbf{\hat{o}}_i) + \texttt{L1}(\mathbf{g}_i, \mathbf{\hat{g}}_i),
\end{equation}
where $\texttt{BCE}$ is binary cross-entropy loss, and $\texttt{L1}$ is L1 loss. The terms $\mathbf{\hat{o}}_i$ and $\mathbf{\hat{g}}_i$ correspond to the ground truth of 3D occupancy and TSDF in level $i$.

The LLM loss, $\mathcal{L}_{llm}$, captures the linguistic aspects by incorporating cross-entropy loss $\texttt{CE}$ for the token identifier of each word:
\vspace{-8pt}
\begin{equation}
\vspace{-8pt}
    \mathcal{L}_{llm} = \texttt{CE}(\mathbf{w}, \mathbf{\hat{w}}),
\end{equation}
where $\mathbf{w}$ denotes to the token identifier of the word in $\mathbf{q}$ and $\mathbf{a}$, and $\mathbf{\hat{w}}$ represents the corresponding ground truth.

\section{Experiments}
In this section, we conduct experiments on different tasks, including 3D reconstruction (see Section~\ref{sec:recon_results}) and 3D vision-language understanding (discussed in Section~\ref{sec:results_scanqa} and Section~\ref{sec:results_3dmv}). We compare our method with the state-of-the-art methods on 3D vision-language understanding and provide an ablation study (see Section~\ref{sec:ablation}) to highlight the effectiveness of each component.

\subsection{Experiment Setup}

\noindent \textbf{Datasets.}
Our setting exclusively accepts video as input, requiring experiments to be conducted on datasets that offer video or multi-view images. Therefore, datasets featuring only point clouds as inputs, such as the Held-In Dataset in 3D-LLM~\cite{hong20233d2}, are unsuitable for our experimentation. Hence, we select the (1) \textbf{ScanNet}~\cite{dai2017scannet} dataset for 3D reconstruction analysis. ScanNet includes 1,513 indoor scenes with videos, depths, and meshes for reconstruction, and we follow the training/validation splits of Atlas~\cite{murez2020atlas}.
We also select two representative 3D vision-language datasets:  (2) the \textbf{ScanQA} dataset~\cite{azuma2022scanqa} contains 41,363 questions from 800 indoor scenes.
(3) the \textbf{3DMV-VQA}~\cite{hong20233d2} dataset that encompasses 50k questions. The questions of 3DMV-VQA are designed to test models on various aspects like concepts, counting, relational, and comparative analyses.
In summary, our dataset selections provide a well-rounded basis for evaluating the effectiveness of our method in both 3D reconstruction and 3D vision-language understanding capabilities.

\noindent \textbf{Implementation Details.}
We use pre-trained image encoders from SAM~\cite{kirillov2023segment} and CLIP~\cite{radford2021learning} as the foundational modules for our 2D encoding process. The dimensions of $\mathbf{e}_\text{sam}$ are specified as 256 channels, with a height of 64 and a width of 64. Similarly, $\mathbf{e}_\text{clip}$ is characterized by 1408 channels, 36 in height, and 36 in width.
The level number $n$ of our 3D decoder is 3, with an input feature of 128 channels. For the ScanNet dataset, the input scales at each level of $\mathbf{e}^{in}$ are set to $30\times40$, $60\times80$, and $120\times160$, while for 3DMV-VQA, the corresponding scales are $32\times32$, $64\times64$, and $128\times128$.
The output scales at each level of $\mathbf{d}$ are defined as $24^3$, $48^3$, and $96^3$ for all datasets.
All scenes are divided into fragments with 9 views each, and these fragments are then reconstructed sequentially, one after another.
Regarding the LLM, we leverage pre-trained QFormer and FlanT5 sourced from BLIP-2~\cite{li2023blip2} to establish the initial weights. The numbers of points and channels of QFormer input $\mathbf{r}_{llm}$ are 10,000 and 1408, respectively.

Our training strategy unfolds in two distinct stages. Initially, the 2D encoder is frozen while the 3D decoder is trained to generate geometric and semantically rich 3D representations and achieve 3D reconstruction. This training phase spans 50 epochs with an initial learning rate of 5e-3, concluding at a final learning rate of 1e-5, and adopts a cosine learning rate scheduler. The training is executed with a batch size of 2 fragments across 16 A100 GPUs. We train the 3D decoder for ScanNet and 3DMV-VQA within 23.5 hours and 29 hours, respectively.

In the subsequent stage, the 2D encoder, 3D decoder, and the majority of the FlanT5 components remain frozen, except for the input and output embeddings. This setting is conducive to training the QFormer and LLM to generate 3D scene-aware dialogue. In this phase, we extract 3D representations and reconstruct whole scenes relevant to question-answer pairs. The learning schedule for training the LLM mirrors that of the 3D decoder, with initial and final learning rates set at 1e-4 and 1e-5, respectively. The batch size for this stage is 16 question-answer pairs, and the training is distributed across 16 A100 GPUs. We train LLM for ScanQA and 3DMV-VQA within 12.5 hours and 33 hours, respectively.

\begin{table*}[tb]
    \centering
    \begin{minipage}[t]{1.0\linewidth}
    \centering
    \caption{\small \textbf{Reconstruction results on ScanNet} for 3D reconstruction analysis. We report standard 2D depth metrics defined in Eigen et al.~\cite{eigen2014depth} and the F-Score metric for geometry.
    }
    \resizebox{0.8\textwidth}{!}{
    \small
    \begin{tabular}{l|cccc}
    \toprule
     & Abs Rel $\downarrow$ & Abs Diff $\downarrow$ & RMSE $\downarrow$ & F-Score $\uparrow$ \\
    \hline
    GPMVS~\cite{hou2019multi} & 0.130 & 0.239 & 0.472 & 0.304 \\ 
    MVDepthNet~\cite{wang2018mvdepthnet} & 0.098 & 0.191 & 0.293 & 0.329 \\ 
    DPSNet~\cite{im2019dpsnet} & 0.087 & 0.158 & 0.232 & 0.344 \\
    COLMAP~\cite{schonberger2016pixelwise} & 0.137 & 0.264 & 0.502 & 0.558 \\ 
    NeuralRecon~\cite{sun2021neuralrecon} & 0.065 & 0.106 & 0.195 & 0.562 \\
    \hline
    \rowcolor{violet!10}
    \methodname (Ours) & \textbf{0.060} & \textbf{0.094} & \textbf{0.182} & \textbf{0.580} \\
    \bottomrule
    \end{tabular}
    }
    \label{tab:scannet-2d}
    \end{minipage}
    \vspace{-20pt}
\end{table*}

\subsection{3D Reconstruction on ScanNet}\label{sec:recon_results}

\noindent \textbf{Baselines.}
We choose NeuralRecon~\cite{sun2021neuralrecon} as the baseline for our reconstruction.
NeuralRecon is widely recognized for its simplicity and effectiveness in 3D reconstruction, providing seamless feature fusion through volumetric features.
To create our 3D geometric and semantic representation, we modify the encoder-decoder framework of NeuralRecon. Furthermore, we compare our reconstruction model with several other competitive approaches. 
COLMAP~\cite{schonberger2016pixelwise} designs a pixel-wise view selection mechanism. MVDepthNet~\cite{wang2018mvdepthnet} uses an encoder-decoder structure with geometric data augmentation. GPMVS~\cite{hou2019multi} uses temporal information between frames. DPSNet~\cite{im2019dpsnet} uses the traditional plane sweep algorithm for dense depth reconstruction.


\noindent \textbf{Metrics.} We utilize the metrics introduced in Atlas~\cite{murez2020atlas} to evaluate our reconstruction results, including Abs Rel, Abs Diff, and RMSE for depth and F-Score for geometry. These metrics represent the relative error, absolute error, and mean-square error when comparing the depth rendered from predicted geometry to the ground truth depth. Additionally, the F-Score metric denotes to the F1 score calculated for the the predicted point cloud in comparison to the ground truth point cloud.

\begin{figure}[!tb]
    \centering
    
 \includegraphics[width=1.0\linewidth]{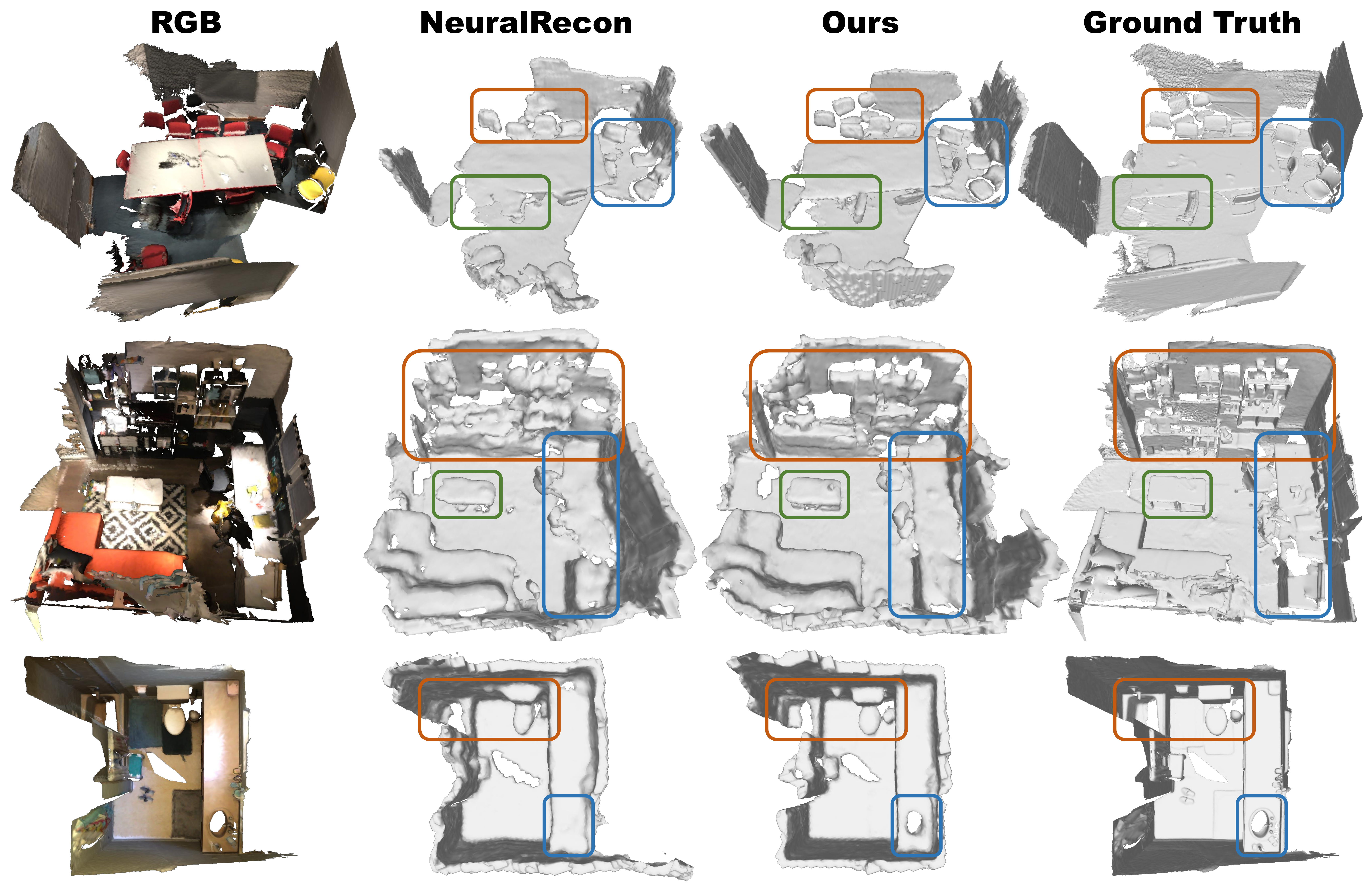}
    \caption{
    \small
    \textbf{3D reconstruction visualization results on ScanNet.} Compared to the baseline method~\cite{sun2021neuralrecon} (second column), our method (third column) predicts the reconstruction results with more semantic details. (Zoom in for details.). 
    }
    \vspace{-20pt}
    \label{fig:recon_vis}
\end{figure}

\noindent \textbf{Results.} As shown in \cref{tab:scannet-2d}, the results highlight the superior reconstruction performance of our \methodname, achieving notable improvements on ScanNet across all metrics compared to other competitive methods. In comparison to the baseline NeuralRecon, the 2D features extracted by our frozen encoder contribute to the 3D geometric and semantic representations in the decoder. Our 3D representations not only encompass more comprehensive information but also exhibit a stronger reconstruction capability, leading to a 1.8$\%$ increase in F-Score.


\noindent \textbf{Qualitative 3D Reconstruction.}\label{sec:recon_vis} \cref{fig:recon_vis} presents the 3D reconstruction visualization results on ScanNet. Thanks to our geometric and semantic representations, \methodname excels in reconstructing objects with high completeness and more intricate edge details. This is evident in finer aspects, such as the back of a chair (top row), the deck (second row), and the wash basin (third row). Our reconstructed objects exhibit detailed coherence without any interruptions or breaks. 

\begin{table}[t]
    \centering
    \begin{minipage}[t]{1.0\linewidth}
    \caption{ \small
    \textbf{Experimental results ($\%$) on ScanQA \textit{val} set.} B-1, B-4 denote BLEU-1, BLEU-4 respectively. Note that our geometry is predicted (pred), while others rely on ground truth (GT).}
    \resizebox{1.0\textwidth}{!}{
    \small
    \setlength{\tabcolsep}{2pt}
    \begin{tabular}{l|c|llcccccc}
    \toprule
        ~ & Recon. & B-1 & B-4 & METEOR & ROUGE-L & CIDER & EM \\  \hline
        
        SingleImage (Flamingo) & - & 23.8 & 8.5 & 10.7 & 29.6 & 52 & 16.9 \\
        MultiView (Flamingo) & - & 25.6 & 8.4 & 11.3 & 31.1 & 55 & 18.8 \\
        SingleImage (FlanT5) & - & 28.6 & 5.1 & 10.6 & 25.8 & 42.6 & 13.3\\
        MultiView (FlanT5) & - & 29.7 & 5.9 & 11.3 & 26.6 & 45.7 & 13.6\\ \hline
        
        VoteNet~\cite{qi2019deep}+MCAN* & GT & 28.0 & 6.2 & 11.4 & 29.8 & 54.7 & 17.3 \\ 
        ScanRefer~\cite{chen2020scanrefer}+MCAN* & GT & 26.9 & 7.9 & 11.5 & 30.0 & 55.4 & 18.6 \\ 
        ScanQA~\cite{azuma2022scanqa}* & GT & 30.2 & 10.1 & 13.1 & 33.3 & 64.9 & \textbf{21.0} \\
        3D-LLM (Flamingo)~\cite{hong20233d}  & GT & 30.3 & 7.2 & 12.2 & 32.3 & 59.2 & 20.4 \\ 
        
        3D-LLM (OPT)~\cite{hong20233d} & GT & 35.9 & 9.4 & 13.8 & 34.0 & 63.8 & 19.3 \\ 
        3D-LLM (FlanT5)~\cite{hong20233d}  & GT & 39.3 & 12.0 & 14.5 & 35.7 & 69.4 & 20.5 \\
        \hline
        3D-LLM (OPT)~\cite{hong20233d} & pred & 32.4 & 8.3 & 12.4 & 31.5 & 60.1 & 14.7 \\
        \rowcolor{violet!10}
        \methodname-LLM (OPT) & pred & 36.6 & 11.0 & 14.0 & 34.4 & 64.3 & 15.8\\
        \hline
        3D-LLM (FlanT5)~\cite{hong20233d} & pred & 35.8 & 9.7 & 13.2 & 32.8 & 61.9 & 15.2 \\
        \rowcolor{violet!10}
        \methodname-LLM (FlanT5) & pred & \textbf{39.8} & \textbf{12.2} & \textbf{14.9} & \textbf{36.3} & \textbf{70.3} & 17.3 \\
        \bottomrule
        
    \end{tabular}}
    \label{tab:scanqa-val}
    \end{minipage}
\end{table}

\subsection{Vision-Language Understanding on ScanQA}\label{sec:results_scanqa}

\noindent \textbf{Baselines.}
We adopt the state-of-the-art method 3D-LLM as our base model for 3D Vision-Language Understanding. However, a direct comparison between our \methodname-LLM and 3D-LLM which uses GT point clouds is not fair. Therefore, we replace the GT point clouds in 3D-LLM with the point clouds predicted by our \methodname to establish a fair baseline. Furthermore, we compare our \methodname-LLM with several other competitive approaches. 
VoteNet+MCAN~\cite{qi2019deep,yu2019deep} applies VoteNet to extract 3D point cloud objects and uses extracted features in the MCAN model for VQA. ScaneRefer+MCAN~\cite{chen2020scanrefer,yu2019deep} employs PointNet++~\cite{qi2017pointnet++} to extract point cloud and lifting features by MCAN. ScanQA~\cite{azuma2022scanqa} fuses the point cloud features with pre-trained language embeddings such as GloVe~\cite{pennington2014glove}. Additionally, we compare our \methodname-LLM with LLMs that take single/multi-view image features as input.


\noindent \textbf{Metrics.} 
We utilize a comprehensive set of metrics to assess performance, including BLEU, METEOR, ROUGE-L, CIDEr, and EM. BLEU (Bilingual Evaluation Understudy) serves as an indicator of text generation accuracy by comparing the overlap of n-grams (sequential word or character sequences) in the generated text with those in reference text. METEOR (Metric for Evaluation of Translation with Explicit Ordering) offers a nuanced evaluation of translation quality, taking into account factors such as precision, recall, stemming, and synonymy. ROUGE-L focuses specifically on measuring the overlap of the longest common subsequences between a reference text (ground truth) and a system-generated summary or text. CIDEr evaluates both word accuracy and the consensus among multiple human-generated reference captions. Lastly, EM (Exact Match) provides a stringent measure of accuracy, quantifying the percentage of predictions that precisely match the ground truth or expected correct answer.

\noindent \textbf{Results.} The effectiveness of our proposed \methodname-LLM on the ScanQA \textit{val} and \textit{test} set is demonstrated in \cref{tab:scanqa-val} and \cref{tab:scanqa-test}. Thanks to our unified representation and reconstruction design, our \methodname-LLM outperforms the baseline 3D-LLM~\cite{hong20233d}---which uses predicted point clouds---by 4.0\% and 4.2\% in BLEU-1 on \textit{val} and \textit{test} of ScanQA. Furthermore, it surpasses all the competitive methods, including 3D-LLM which uses GT point clouds, across the most of metrics on both the ScanQA \textit{val} and \textit{test} set.

\begin{table*}[t]
    \centering
    \begin{minipage}[t]{1.0\linewidth}
    \centering
    \small
    \caption{\small \textbf{Experimental results ($\%$) on ScanQA \textit{test} set.} B-1, B-4 denote BLEU-1, BLEU-4 respectively. Note that our geometry is predicted (pred), while others rely on ground truth (GT). (*: using explicit object representations). }
    \vspace{-10pt}
    \resizebox{0.95\textwidth}{!}{
    \small
    \setlength{\tabcolsep}{2pt}
    \begin{tabular}{l|c|cccccc}
        \toprule
        ~ & Recon. & B-1 & B-4 & METEOR & ROUGE-L & CIDER & EM \\ \hline
        VoteNet~\cite{qi2019deep}+MCAN* & GT & 29.5 & 6.0 & 12.0 & 30.9 & 58.2 & 19.7 \\ 
        ScanRefer~\cite{chen2020scanrefer}+MCAN* & GT & 27.9 & 7.5 & 11.9 & 30.7 & 57.4 & 20.6 \\ 
        ScanQA~\cite{azuma2022scanqa}* & GT & 31.6 & \textbf{12.0} & 13.5 & 34.3 & 67.3 & \textbf{23.5} \\
        3D-LLM (Flamingo)~\cite{hong20233d} & GT & 32.6 & 8.4 & 13.5 & 34.8 & 65.6 & 23.2 \\ 
        
        3D-LLM (OPT)~\cite{hong20233d} & GT & 37.3 & 10.7 & 14.3 & 34.5 & 67.1 & 19.1 \\ 
        3D-LLM (FlanT5)~\cite{hong20233d} & GT & 38.3 & 11.6 & 14.9 & 35.3 & 69.6 & 19.1 \\
        \hline
        3D-LLM (OPT)~\cite{hong20233d} & pred & 33.0 & 8.9 & 12.6 & 31.7 & 60.9 & 14.8\\
        \rowcolor{violet!10}
        \methodname-LLM (OPT) & pred & 38.4 & 10.9 & 14.7 & 34.8 & 68.2 & 17.2\\
        \hline
        3D-LLM (FlanT5)~\cite{hong20233d} & pred & 35.5 & 9.6 & 13.0 & 32.2 & 62.1 & 14.9\\
        \rowcolor{violet!10}
        \methodname-LLM (FlanT5) & pred & \textbf{39.7} & \textbf{11.9} & \textbf{15.3} & \textbf{36.6} & \textbf{71.7} & \textbf{19.3} \\
        \bottomrule
    \end{tabular}}
    \label{tab:scanqa-test}
    \end{minipage}
    \vspace{-10pt}
\end{table*}

\begin{table*}[t]
    \centering
    \begin{minipage}[t]{1.0\linewidth}
    \centering
    \caption{\small \textbf{Experimental results ($\%$) on 3DMV-VQA~\textit{test} set} with four question types: Concept (Conc), Counting (Coun), Relation (Rela), and Comparison (Comp). Note that our geometry is predicted (pred), while others rely on ground truth (GT) except SingleImage and MultiView. (*: using explicit object representations).
    }
    \vspace{-10pt}
    \resizebox{0.8\textwidth}{!}{
    \setlength{\tabcolsep}{2pt}
    \begin{tabular}{l|c|ccccc}
    \toprule
    & Recon. & Conc & Coun & Rela & Comp & Overall\\ \hline
    SingleImage (Flamingo) & - & 58.7 & 18.5 & 38.4 & 60.1 & 40.3\\
    MultiView (Flamingo) & - & 60.0 & 18.3 & 40.2 & 61.4 & 41.6\\
    SingleImage (FlanT5) & - & 58.0 & 20.4 & 42.3 & 62.3 & 43.1\\
    MultiView (FlanT5) & - & 61.9 & 21.1 & 48.0 & 62.3 & 47.1\\
    \hline 
    NS-VQA*~\cite{yi2018neural} & GT & 59.8 & 21.5 & 33.4 & 61.6 & 38.0 \\ 
    3D-CLR*~\cite{hong20233d2}  & GT & 66.1 & \textbf{41.3} & 57.6 & 72.3 & 57.7 \\ 
    3D-LLM (Flamingo)~\cite{hong20233d} & GT & 68.9 & 32.4 & 61.6 & 68.3 & 58.6 \\
    
    3D-LLM (OPT)~\cite{hong20233d} & GT & 63.4 & 30.7 &  57.6 & 65.2 & 54.9 \\
    3D-LLM (FlanT5)~\cite{hong20233d} & GT & 68.1 & 31.4 & 55.1 & 69.7 & 54.6\\
    \hline
    3D-LLM (OPT)~\cite{hong20233d} & pred & 58.2 & 25.8 &  53.2 & 59.1 & 51.8 \\
    \rowcolor{violet!10}
    \methodname-LLM (OPT) & pred & 67.4 & 27.3 & 59.0 & 69.5 & 58.0 \\
    \hline
    3D-LLM (FlanT5)~\cite{hong20233d} & pred & 61.3 & 27.1 & 55.1 & 62.3 & 53.9\\
    \rowcolor{violet!10}
    \methodname-LLM (FlanT5) & pred & \textbf{70.8} & 28.6 & \textbf{62.8} & \textbf{77.7} & \textbf{62.0} \\
    \bottomrule 
    \end{tabular}}
    \label{tab:3dmv-vqa}
    \end{minipage}
    \vspace{-15pt}
\end{table*}

\begin{figure*}[t]
    \centering
    \includegraphics[width=1.0\linewidth]{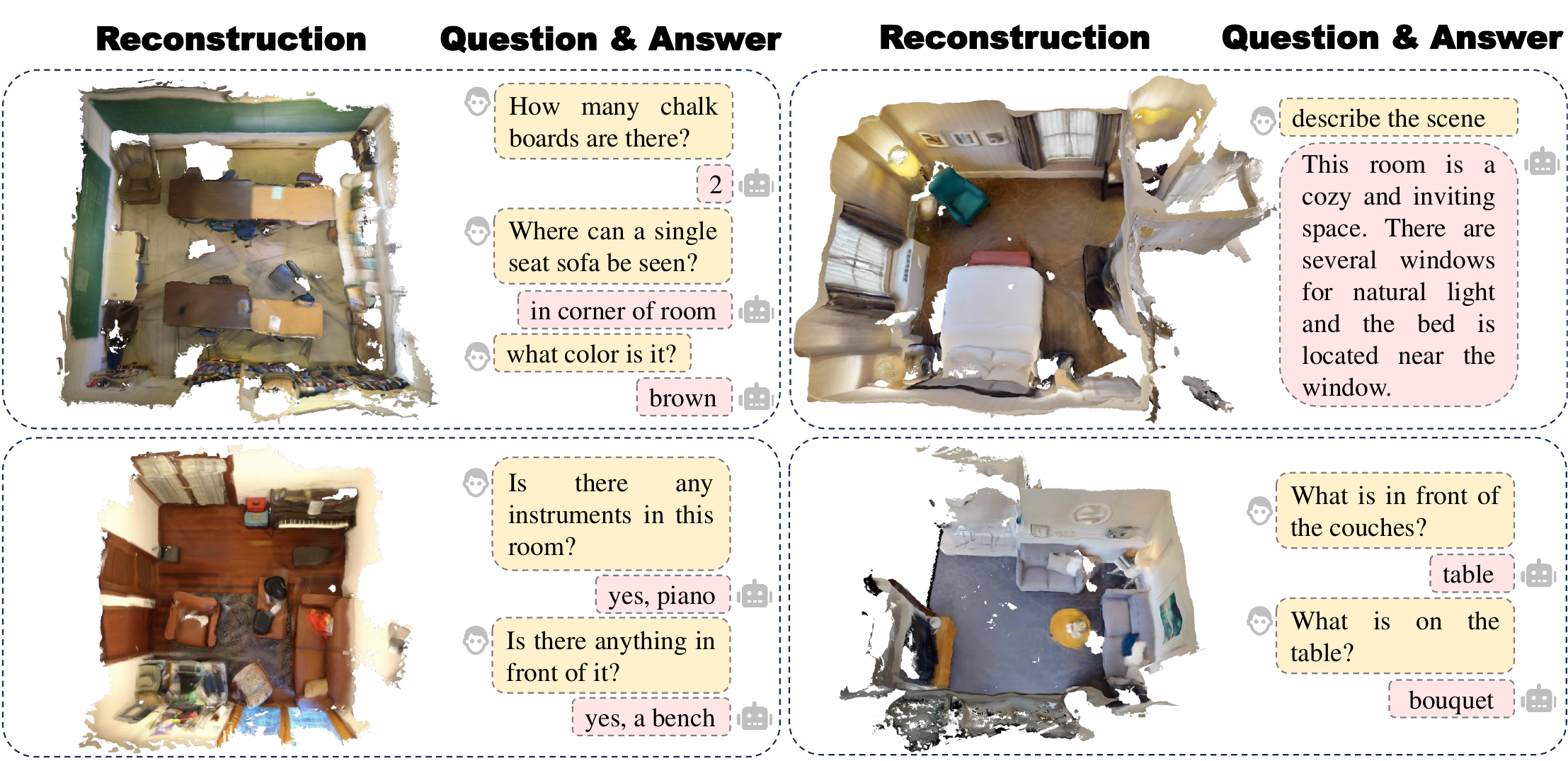}
    \caption{
    \small
    \textbf{3D vision-language understanding visualization results.} Our \methodname-LLM predicts accurate reconstructions and answers the user input questions. (Zoom in for details.).
    }
    \label{fig:qa_vis}
\end{figure*}

\subsection{Vision-Language Understanding on 3DMV-VQA}\label{sec:results_3dmv}

\noindent \textbf{Baselines.}
We compare \methodname-LLM with several competitive approaches. NS-VQA~\cite{yi2018neural} takes 2D grounding features and prior symbolic structure as inputs. 3D-CLR~\cite{hong20233d2} associates 3D features with 2D features via a 3D-2D alignment loss. And the state-of-the-art 3D-LLM~\cite{hong20233d} which uses GT point clouds.


\noindent \textbf{Results.}
Thanks to our unified 3D representation and reconstruction design, our \methodname-LLM achieves the highest overall accuracy at 62.0$\%$ with the predicted reconstruction results instead of relying on GT point clouds. It also leads in the individual categories of Concept, Relation, and Comparison with 70.8$\%$, 62.8$\%$, and 77.7$\%$ respectively.

\begin{table}[t]
    \centering
    \begin{minipage}[b]{1.0\linewidth}
    \centering
    \caption{\small \textbf{Ablation studies about encoder design choices on ScanNet} for reconstruction analysis.
    }
    \vspace{-10pt}
    \resizebox{0.8\textwidth}{!}{
    \centering
    \small
    \begin{tabular}{ccc|cccc}
    \toprule
    Recon. & CLIP & SAM & Abs Rel $\downarrow$ & Abs Diff $\downarrow$ & RMSE $\downarrow$ & F-Score $\uparrow$ \\
    \hline
    NeuralRecon & \xmark & \xmark & 0.065 & 0.106 & 0.195 & 0.562 \\
    $\text{Rec}_\text{C}$ & \cmark & \xmark & 0.069 & 0.113 & 0.213 & 0.534 \\
    $\text{Rec}_\text{S}$ & \xmark & \cmark & 0.070 & 0.106 & 0.205 & 0.570 \\
    \rowcolor{violet!10}
    $\text{Rec}_\text{CS}$ & \cmark & \cmark & \textbf{0.060} & \textbf{0.094} & \textbf{0.182} & \textbf{0.580} \\
    \bottomrule
    \end{tabular}}
    \label{tab:appendix_encoder}
    \end{minipage}
    \vspace{-5pt}
\end{table}

\begin{table}[t]
    \centering
    \begin{minipage}[b]{1.0\linewidth}
    \centering
    \caption{ 
    \textbf{Ablation studies on ScanQA \textit{val} set} about representation~(Repre.) and reconstruction~(Recon.).}
    \vspace{-10pt}
    \resizebox{0.9\textwidth}{!}{
    \begin{tabular}{l|c|cccccccc}
    \toprule
    Repre. & Recon. & B-1 & B-4 & METEOR & ROUGE-L & CIDER & EM \\ \hline
    CLIP (MultiView) & - & 29.7 & 5.9 & 11.3 & 26.6 & 45.7 & 13.6\\ \hline
    $\text{CLIP}_\text{proj}$ (3D-LLM) & GT & 39.3 & 12.0 & 14.5 & 35.7 & 69.4 & \textbf{20.5} \\ \hline
    $\text{CLIP}_\text{proj}$ (3D-LLM) & $\text{Rec}_\text{CS}$ & 35.8 & 9.7 & 13.2 & 32.8 & 61.9 & 15.2 \\ 
    3D Feat (based on CLIP) & $\text{Rec}_\text{C}$ & 37.8 & 11.2 & 14.3 & 35.6 & 67.9 & 17.8 \\
    3D Feat (based on SAM) & $\text{Rec}_\text{S}$ & 37.9 & 10.2 & 14.3 & 35.7 & 68.1 & 18.3 \\
    \rowcolor{violet!10}
    3D Feat (\methodname-LLM) & $\text{Rec}_\text{CS}$ & \textbf{39.8} & \textbf{12.2} & \textbf{14.9} & \textbf{36.3} & \textbf{70.3} & 17.3 \\
    \bottomrule
    \end{tabular}}
    \label{tab:ablation_scanqa}
    \end{minipage}
    \vspace{-15pt}
 \end{table}

\subsection{Qualitative 3D Vision-Language Understanding.}
We also provide the qualitative results about 3D vision-language understanding in \cref{fig:qa_vis}. Our unified 3D representation and reconstruction capabilities empower the LLM to accurately answer questions necessitating a comprehensive understanding of objects and their spatial locations in 3D scenes. Our method provides accurate answers to questions about color, object numbers, and locations, thanks to our semantically rich representation.

\subsection{Ablation Studies}\label{sec:ablation}
\noindent \textbf{Encoder Design Choices.}
The encoder in Our \methodname incorporates two frozen pre-trained models: the CLIP visual encoder and the SAM visual encoder. In \cref{tab:appendix_encoder}, we detail the results of our ablation studies, which quantify the contribution of each component to the 3D reconstruction.
The study compares four configurations: ``NeuralRecon'' without either pretrained encoder, ``$\text{Rec}_\text{C}$'' with only the frozen CLIP encoder, ``$\text{Rec}_\text{S}$'' with only the frozen SAM encoder, and ``$\text{Rec}_\text{CS}$'' with both encoders active.
A comparison between the performance of ``$\text{Rec}_\text{C}$'' and ``NeuralRecon'' indicates that Clip features, which focus on semantics, display reduced performance while still providing valuable geometric information for 3D reconstruction. When Comparing ``$\text{Rec}_\text{S}$'' to ``NeuralRecon'', it becomes apparent that SAM features, enriched with object-level information, demonstrate superior geometric capabilities compared to the baseline model with a trained encoder. Moreover, an analysis of the performance of ``$\text{Rec}_\text{CS}$'' in relation to both ``$\text{Rec}_\text{C}$'' and ``$\text{Rec}_\text{S}$'' reveals that the fusion of Clip features and SAM features yields more comprehensive information, resulting in superior reconstruction outcomes.

\noindent \textbf{Representation.}
In \cref{tab:ablation_scanqa}, we demonstrate the importance of representation on the ScanQA dataset.
The study compares six configurations: ``CLIP (MultiView)'' without lifting and geometry, 3D-LLM denoted as ``CLIP$_\text{proj}$ (3D-LLM)'' with ``GT'' point clouds, the baseline represented by ``CLIP$_\text{proj}$ (3D-LLM)'' with point clouds predicted by ``$\text{Rec}_\text{CS}$'', ``3D Feat'' involving unified representation and reconstruction of ``$\text{Rec}_\text{C}$'', ``$\text{Rec}_\text{S}$'', and ``$\text{Rec}_\text{CS}$''.
Comparing the performance of ``CLIP (MultiView)'' and the baseline underscores the significance of geometry in 3D LLM. The evaluation of the baseline against all configurations of ``3D Feat'', unified representation, and reconstruction significantly enhances performance in 3D vision-language understanding. And our final \methodname-LLM demonstrates a notable improvement of 4.0\% and 4.2\% in BLEU-1 on \textit{val} and \textit{test} of ScanQA compared to the baseline. Comparing \methodname-LLM with both ``3D Feat (based on CLIP)'' and ``3D Feat (based on SAM)'', the fusion of two features demonstrates an enhancement in LLM capabilities. Furthermore, a comparison between the baseline and 3D-LLM reveals a significant influence on reconstruction quality. Finally,  when comparing our \methodname-LLM with 3D-LLM, our unified model effectively mitigate the impact of reconstruction quality, resulting in higher performance compared to 3D-LLM which uses GT point clouds.





\section{Conclusion}
Our research addresses a critical challenge in applying Large Language Models (LLMs) to 3D environments: the acquisition of effective 3D representations suitable for LLMs. Compared to structural 1D text or 2D images, 3D scene representations are hard to interpret and manipulate, which highlights the need for research of advanced processing and representation learning algorithms. We emphasize the significance of geometric and semantically rich features and introduce a unified module for 3D representation and reconstruction.
Future work will explore scaling our method to enhance more 3D capabilities with LLMs, including advancing 3D scene perception and 3D generation.



\section*{Acknowledgement}
This work is supported by Guangdong Basic and Applied Basic Research Foundation (Grant No.2024A1515012043), the National Key R \& D Program of China (2022ZD0160201), and Shanghai Artificial lntelligence Laboratory.

%
%

\bibliographystyle{splncs04}

\end{document}